\documentclass{article}
\usepackage[utf8]{inputenc}
\usepackage{authblk}
\usepackage{setspace}
\usepackage[margin=1.25in]{geometry}
\usepackage{graphicx, color}
\graphicspath{ {./figures/} }
\usepackage{subcaption}
\usepackage{amsmath}
\usepackage{lineno}

\title{On a Functional Definition of Intelligence}

\author[1*$\dag$]{Warisa Sritriratanarak}
\author[2$\dag$]{Paulo Garcia}

\affil[1]{Computer Engineering, Chulalongkorn University, Bangkok, Thailand.}
\affil[2]{AI \& Robotics Program: International School of Engineering, Chulalongkorn University, Bangkok, Thailand.}
\affil[*]{Address correspondence to: Paulo.G@chula.ac.th}
\affil[$\dag$]{These authors contributed equally to this work.}

\date{}

\begin{document}

\maketitle

\begin{abstract}
Without an agreed-upon definition of intelligence, asking “is this system intelligent?” is an untestable question. This lack of consensus hinders research, and public perception, on Artificial Intelligence (AI), particularly since the rise of generative- and large-language models. Most work on precisely capturing what we mean by “intelligence” has come from the fields of philosophy, psychology, and cognitive science. Because these perspectives are intrinsically linked to intelligence as it is demonstrated by natural creatures, we argue such fields cannot, and will not, provide a sufficiently rigorous definition that can be applied to artificial means. Thus, we present an argument for a purely functional, black-box definition of intelligence, distinct from how that intelligence is actually achieved; focusing on the “what”, rather than the “how”. To achieve this, we first distinguish other related concepts (sentience, sensation, agency, etc.) from the notion of intelligence, particularly identifying how these concepts pertain to artificial intelligent systems. As a result, we achieve a formal definition of intelligence that is conceptually testable from only external observation, that suggests intelligence is a continuous variable. We conclude by identifying challenges that still remain towards quantifiable measurement. This work provides a useful perspective for both the development of AI, and for public perception of the capabilities and risks of AI.
\end{abstract}


\section{Introduction}

    The rise in power and popularity of recent machine learning technologies (particularly, Large Language Models (LLMs) such as ChatGPT \cite{doi:10.1126/sciadv.adh1850}) have fueled the debate on intelligence, as well as on the dangers of sufficiently powerful Artificial Intelligence (AI) \cite{doi:10.1126/science.aag3311}. Considerations on Ethical AI, AI alignment, and related concepts, are now part of popular discourse \cite{doi:10.1126/science.aay5189}.
    \par These discussions, as well as more fundamental debates about the differences between AI, Artificial General Intelligence (AGI), and super-intelligence, have been hindered (and, in the authors' opinion, fundamentally hampered within the traditional definition of scientific theory as per Popper \cite{donagan1964historical}) by an imprecise, or even completely absent, definition of \textit{intelligence} \cite{legg2007collection}, as well as by a lack of orthogonality between such definitions and definitions of other anthropomorphic concepts such as autonomy, conscience, and emotion \cite{briffa2022should}. Such a lack of definition results in discourses where multiple parties hold fairly distinct, internally incomplete and inconsistent definitions of intelligence, making all efforts to categorize different forms of AI useless. Thus, we observe prominent questions such as "is ChatGPT intelligent?" \cite{wang2023does}, which are nonsensical (and completely untestable, from a scientific perspective) without strict definitions, and agreed-upon evaluation criteria.
    \par This motivation fuels the arguments in this paper: our objective is to (attempt to) provide a strict definition of intelligence, free from linguistic or cultural ambiguity, that can be tested objectively. Thus, the inclusion of this research question ("How to define intelligence and establish the evaluation and standardization framework for intelligent computing?") as part of the 10 Fundamental Scientific Questions on Intelligent Computing \cite{10measures}.

    \par Our thesis, that frames most of our arguments, is that such a definition must be constructed by mathematical principles, and phrased exclusively on externally visible properties: system inputs and outputs, as a functional definition of intelligence. Historically, attempts to capture the mechanism of intelligence have been pursued by philosophers \cite{clark2000mindware}, cognitive scientists \cite{laird2017standard}, and the like. These efforts are intrinsically linked to a specific instance of intelligence (human and animal intelligence) and are not necessarily appropriate to capture artificial instances, much like bio- or zoological definitions of swimming are not appropriate to study the motion of submarines (paraphrasing Dijkstra \cite{submarine}). Thus, we pursue a functional definition and evaluation, which is independent of the specific instance of intelligence performing the reasoning.
    \par Fundamentally, this article presents descriptive research that formulates a quantifiable, testable, measure of intelligence. The goal is to equip researchers with objective tools to measure intelligence, and, more critically, to achieve a precise definition that can inform the development of AI. We pursue this by first providing an argument for what intelligence is \textit{not} (but is often confused with) in Section \ref{sec:methods}, including properties that often exist in artificial systems. We then provide an argument for what intelligence \textit{is}, and we formalize that definition mathematically in Section \ref{sec:results}. We conclude by discussing how to test that measure in Section \ref{sec:discussion}.

\section{Materials and Methods}\label{sec:methods}

    \begin{figure}[t]
    \centering
        \includegraphics[width=0.5\columnwidth]{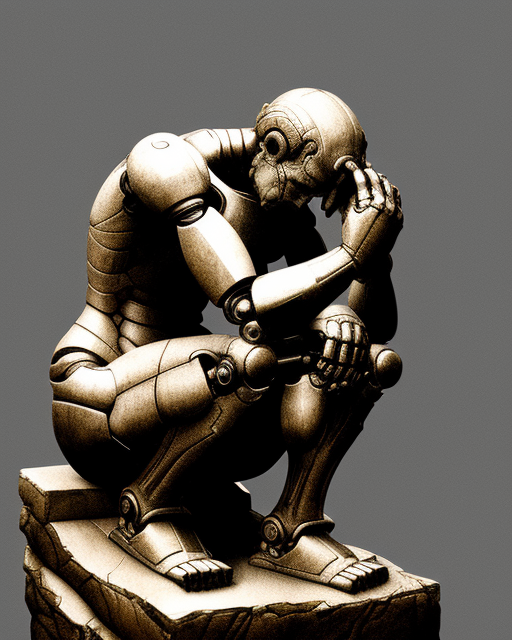}
        \caption{Image generated by "starryai" through the prompt "A robot sitting in the position of “the thinker by rodin”, digital painting,  digital illustration,  extreme detail,  digital art,  4k,  ultra hd"}
        \label{fig:robot}
    \end{figure}

    Our methodological approach is argumentative \cite{lavery2020argumentation}. Whilst this approach is rarely used in science and engineering, where empirical and/or formal methods are the \textit{de facto} methodologies, we believe it is the right approach to tackle this problem, since, fundamentally, the goal is to achieve a strict definition. It is not possible, to the best of the authors' knowledge, to empirically or formally verify a definition. The reverse is true: it is necessary to have an agreed upon definition to be able to pursue those methodologies \cite{abbott2002definitions}. Thus, we provide a logical argument that captures a testable definition of intelligence towards eventually being able to perform this kind of research.
    \par We begin by exploring what intelligence is not, as part of our argumentative methodology, in the subsections below: we then present our definition of intelligence as a logical sequitur to these arguments. 

\subsection{Observability}

    A fundamental concept in testing engineering is the distinction between white-box and black-box testing (including shades of grey-box testing in between) \cite{khan2012comparative}. In \textit{black-box testing}, the internals of the artifact under test are not observable. Rather, only its interfaces with the outside world are: testing provides inputs, and observes whether outputs correspond to expectations (across several metrics, e.g., correctness, response time). This testing modality is particularly useful when the agents performing the tests are not the ones that developed the artifact: the goal is to hide implementation details, such that testers are not biased by the implementation and focus instead on testing inputs that are representative of the total possible input space. 
    \par In \textit{white-box testing}, internals are fully visible. This is particularly important when an artifact has failed black-box testing: white-box testing allows developers to reproduce the failing tests, observing intermediate steps between input and output, to determine where and how implementation fails.

    \par We advocate that a useful definition of intelligence must be testable through black-box testing (Fig. \ref{fig:robot}). As per our previous statement, we believe a definition should be independent of the specific instance of intelligence performing the reasoning. Thus, our focus on functional definitions: a system should be deemed intelligent strictly by its response, given inputs. This prevents refutation of intelligence based on process; e.g., "that's not how humans think" \cite{wing2006computational}; and is aligned with the intention, if not the specific methods of, established tests such as Turing's \cite{Yampolskiy2013}.
    \par An important caveat must be addressed. There should be a discussion on the distinction between intelligence, and faking intelligence \cite{srivastava2022beyond}.  
    
    \par There is a non-insignificant number of people who believe ChatGPT (and other contemporary LLMs) passes the Turing test \cite{noever2022turing}. This has led to accusations of moving the goal posts, when professing that evaluating LLMs on the Turing test doesn't suffice to qualify them as intelligent (for the record, the authors' opinion is indeed that ChatGPT and other LLMs are \textit{somewhat} intelligent, at least for a full definition; more on that later).
    \par Turing did not foresee the stochastic power of computations and data sizes at modern scale. Thus, his test (whose intention we agree with, that of black-box testing) is insufficiently formulated, because it incorrectly assumed that sufficiently high skill at the task of conversation implied intelligence. We believe refuting the Turing test as a valid metric is not moving the goal posts, but rather realizing that there are important aspects unconsidered in the original formulation, that render the original test meaningless from the beginning.
    \par More recent formulations on intelligence and corresponding tests \cite{chollet2019measure}, published \textit{before} the widespread success of LLMs, have already shown that arbitrary skill level on any one application is an insufficient measure of intelligence. LLMs are impressively capable of demonstrating skill within language use, in the same way chess engines are impressively capable of demonstrating skill within chess. Is this intelligence? We argue that \textit{yes}.... partially.

\subsection{Orthogonal concepts}

    Before we attempt to define intelligence, let us examine a few related concepts that are often (incorrectly) bundled together with intelligence, especially from a perspective of unnecessary anthropomorphization \cite{yampolskiy2013artificial}. Our goal here is to demonstrate that these concepts are orthogonal to intelligence (in fact, most of them can be observed and measured in systems without any intelligence). By doing so, we are outlining the boundaries that must be obeyed by a definition of intelligence.

    \par It is also important to highlight that much of the discussion on AI safety, ethics, and alignment, is predicated on the presence of several of these other concepts. For example, AI safety \cite{amodei2016concrete} is considerably different for systems with and without autonomy (described below). Whilst this discussion is beyond the scope of this paper, we believe the distinctions presented here can also guide it \cite{charisi2017towards}.

    \begin{itemize}
        \item \textbf{Sensations}. From a purely functional perspective (i.e., what sensations \textit{achieve} by being experienced), sensations can be defined as triggers for specific actions. Consider, for example, acute physical pain: in living beings, it notifies the the brain that a part of the body is being damaged. It serves the evolutionary goal of self-preservation, by identifying attacks on that preservation, allowing the brain to issue commands that initiate corrective actions (e.g., removing the limb from the source of harm). Persistent pain indicates that a part of the body is damaged or vulnerable, and limited action should be performed.
        \par Diagnostic information in artificial systems (e.g., temperature sensors, pressure sensors, software checksums) \cite{kawabata2002study} with corresponding reaction mechanisms perform the same action: it triggers a response that seeks to avoid damage (again, this can and is currently done in several systems, without any form of intelligence; merely automation). 
        \par Similarly, hunger notifies the brain that a system is low on a critical resource, triggering an action to replenish that resource. Battery monitoring, coupled with automated charging (e.g., autonomous robots) performs exactly the same behavior. \textbf{Functionally, artificial sensations have existed for decades}. 
        \par Notice that it is trivial to design experiments that test whether an entity is experiencing pain or hunger, regardless of that entity being natural or artificial, without ever having the need to philosophically define what it means to experience that feeling; it suffices to demonstrate its function at play.

        \item \textbf{Autonomy}. Within engineering, a system is typically deemed autonomous, to a certain degree, if it is capable of effecting actions without external supervision or approval, within that degree \cite{bradshaw2013seven}. For example, a mobile robot is typically considered autonomous if it is capable of moving without external control. Its "desire" to reach a certain position, i.e., its objective, pertains to agency (see below), not autonomy; autonomy is a measure of its ability to effect the actions it has chosen (or have been chosen for it) to achieve those objectives.
        \par In this light, many autonomous systems exist, most of which without any intelligence; from robotic or other mechanical systems capable of initiating motion, to software systems capable of  initiating computation or communication. Again, testing this is trivial: one must merely design an experiment to observe whether an operating artifact effects some operation without requiring external assistance.

        \item \textbf{Agency}. A system is deemed an agent if it has a goal (i.e., an intention, an objective), and is capable of performing decisions towards reaching that goal \cite{floridi2023ai}. A simple control system has agency (i.e., is an agent): it has a goal (making a control variable reach a certain value) and it can perform decisions towards that goal (e.g., the value to set the actuation variable to). A system may have agency and autonomy (e.g., a control system that can directly, autonomously actuate) or agency without autonomy (e.g., a control system that can recommend, but not actuate on, the actuation variable). Agency does not imply intelligence: a system may make nonsensical decisions (because it is not intelligent or skilled), but that still demonstrates agency. 

        \item \textbf{Skill}. Prior work has already clearly outlined the differences between skill and intelligence \cite{chollet2019measure}, but it is useful to provide a short summary here. Skill at a certain task means that a system demonstrates a high level of proficiency at that task. Formally, skill can be defined in the following way: 
        \par Given the set $N$ of all possible inputs to a system when performing a task. For each element $n \in N$, there is a set of possible system outputs $M_n$ that will result in some action on the task at hand. Assuming outputs $M_n$ are ranked (ordered) according to how closely their outcome on the task is to the goal of the agent, a system is deemed skillful if it statistically selects the highest ranked outputs (up to whatever threshold is required) across all values of $N$. As an example, a system is skilled at chess if, for all presented inputs (board positions), it selects a move that will likely lead to victory, over $x\%$ of the time, for whatever value of $x$.

        \par Notice, again, that it is possible for a system to be extremely skilled at a task or a set of tasks, without \textit{any other intelligence whatsoever}. For example, for tasks where the state space is tractable with contemporary computing power and storage, a look-up table can be built for all possible inputs, resulting in a system that selects the best output one hundred percent of the time. What we have observed with LLMs (and other machine learning systems) is that, for many tasks, it is not necessary to know the entire state space \textit{a priori}; a statistical model of the state space suffices to demonstrate high skill level. We will argue that this is indeed \textit{part}, but not all, of intelligence.

        \item \textbf{Sentience}. Consciousness and self-awareness are often used interchangeably with sentience, and we do not identify any distinction that would warrant separate definitions: our belief is that the use of these heterogeneous terms comes either from linguistic history, or from ambiguous definitions in the minds of the utterers \cite{kagan2022vitro}. 
        \par If a system possesses an internal model of the world, and that model includes the system itself, it is sentient. It is "self-aware". Of course, this is not a binary distinction: a robot that models its own position in the world is self-aware, to an extent: but it is possible it does not model its agency, or autonomy, or other properties. Does this make it not "conscious"? We believe saying "no" is, again, unnecessary anthropomorphization. Humans are not, most of the time, aware of several internal properties: heartbeat and breathing are usually in autopilot. The rationale for decisions is not always clear to most humans: thus, the practices of mindfulness and reflection. Much like intelligence, sentience is analog.
   
    \end{itemize}

    \subsection{Components of intelligence}

    Intelligence is a combination of (at least) three properties: the ability to \textit{learn}, the capacity to store \textit{knowledge}, and the ability to \textit{reason}. These are, of course, overloaded terms, so let us attempt a precise definition.

        \subsubsection{Learning}

        \begin{itemize}
        \item \textit{Learning} means that an entity can improve its \textit{knowledge} and \textit{reasoning} (whatever that means), given either new information available, or repeated instances of old information (i.e., repeated experiences) \cite{akyurek2022learning}. Specifically, learning should improve certain aspects of reasoning (the ones related to the information at hand) without affecting aspects of reasoning pertaining to unrelated information. 
        \end{itemize}

        \par If one accepts a definition of reasoning as "classifying inputs into classes" \cite{ripley1994neural} (which we believe is an insufficient definition in the general case, but it suffices to discuss learning when it comes to most neural network architectures), we can reason (no pun intended) about whether or not neural networks learn.
        \par Consider the case where a neural network, regardless of the architecture, is initialized with random weights (i.e., it is untrained). Its reasoning ability (its capability to correctly classify inputs, regardless of the number of classes) is quite poor. As we train the neural network, using established techniques such as gradient descent and back propagation, its classification ability improves. Ergo, according to the provided definition of learning, the network is indeed learning.
        \par This exercise becomes more interesting as we expand the thought experiment to consider a case where a network is deemed "trained": i.e., it achieves classification accuracy above a certain established threshold (e.g., $> 95\%$) for all classes. Furthermore, let us assume this state is pareto-optimal within the space of possible weights. Thus, further training towards improving classification accuracy on one class must necessarily decrease accuracy on at least one other class (otherwise, the initial state would not be pareto-optimal).
        \par In this scenario, is the network still learning?
        \par According to the presented definition, no; new information is hindering reasoning on aspects pertaining to unrelated information. For the network to continue learning, it would be necessary for the number of weights to be increased, such that it becomes possible to find a new state, now in a larger space of possible weights, where accuracy increases across all classes. 
        \par This reasoning implies that learning is possible only if the storage that supports reasoning is arbitrarily large, much in the same way that Turing-machines assume an infinite tape: a real computer is only Turing-equivalent insofar as the required memory for the computations does not exceed physical (bounded) limits. It is perfectly reasonable to assume that any physical entity is a bounded approximation of an abstract learning machine, but this point must be taken into consideration when testing for learning capability: \textbf{testing for learning must be performed at storage requirements below physical limits}, much like testing for computational power must be performed at memory requirements below physical limits \cite{koppula2015indistinguishability}.

        \subsubsection{Knowledge}

        Knowledge is perhaps the easiest of the terms to define: information about the world \cite{halpern1986reasoning}. This includes models of objects and all their associated properties (size, weight, color, etc.); models of classes that objects belong to ("toy", "vehicle", "plant", etc.); models of relationships between classes ("is a", "has", etc.); identification of these relationships as abstract objects ("ownership", "fairness", etc.), into a recursive ontology of many strange loops \cite{hofstadter2007strange}.
        \par Knowledge also includes models of \textit{actions}: functions an object can perform that affect the properties of other objects (i.e., cause and effect relationships), which in turn include triggers (i.e., a change to an object property may trigger an action, and a consequence of an action may be the triggering of other actions, in complicated causal sequences).
        \par It is not lost on us that our definition of knowledge mirrors computing paradigms: our definitions of "models" and "actions" are analogous to "data" and "code" \cite{van2010abstracting}, or "closures" and "functions" \cite{lassez1984closures}.

        \subsubsection{Reasoning}

        Reasoning is arguably the most elusive term, but now that we have defined what it is \textit{not}, perhaps we can approximate a definition using previously defined terms. Let us separate it into two different aspects: \textit{logical reasoning} and \textit{planning}.
        \par Logical reasoning is the ability to infer causes from effect \cite{bronkhorst2020logical}. Given an internal world-model corresponding to an instance of knowledge (read: a set of objects that are instances of models, actions, and respective relationships, as defined in \textit{knowledge}) in a certain state (i.e., with specific values for every known property), denoted as $W_S$; logical reasoning is the ability to accurately predict the world in the next state $W_{S\prime}$ given an action performed inside the world-model. I.e., logical reasoning is the ability to propagate causal effects across chains of objects \cite{kaliszyk2017holstep}. 
        \par One system is capable of "better" logical reasoning than another, if, all other things being equal, the number of correct causal relationships minus the number of incorrect causal relationships in its internal world-model is higher than the other system's (assuming internal rules of logical inference are correct, thus correctly calculating all known causes-effects as per the world-model's relationships \cite{dewall2008evidence}). Thus, we are stating that logical reasoning is nothing more than the capability to apply the rules of formal logic, subject to the scope and correctness of the internal world-model.
        \par \textit{Planning} is the capability of, given an internal world-model in a certain state $W_S$, and a desired world-model in a new state $W_{S\prime}$, choose, in a timely manner, a set of actions that will make the world transition to the desired state: i.e., "what is the best course of action to achieve this goal?" \cite{lithner2017principles}.
        \par Two things must be noted here. First, there is no mention of accuracy regarding the real world; merely, internal consistency. If the internal predictions are incorrect, that is a failure of knowledge, not of reasoning. Second, "in a timely manner" carries a lot of weight in this definition, although performance not typically being associated with reasoning. This is necessary: given correct knowledge and boundless time, a system could select the best set of actions by brute-force alone. How quickly the system arrives at its conclusions must matter to quantify its planning capabilities (even if these are based on heuristics). There is, of course, a tradeoff between speed and accuracy: system A may arrive at a set of actions resulting in $W_A$ faster than system B arrives at a set of actions resulting in $W_B$, where $W_A \neq W_B, W_A \neq W_{S\prime}, W_B \neq W_{S\prime}$, but $|W_A - W_{S\prime}| > |W_B - W_{S\prime}|$: a weighted sum of accuracy and speed must be applied.

\section{Results}\label{sec:results}
With preliminaries achieved, we can now offer a quantifiable measure of intelligence. Let $I_i$ denote the intelligence of system $i$. Let $W^*$ denote the real world, with all its objects and actions, in its current state (to the best of our perception and measurement capabilities) and $W^i_S$ denote the system's internal world-model in state $S$. A world-model $W^i_S$ is defined as tuple consisting of the set of models $M$, the set of actions $A$, the set of concrete objects (models instances) $O$ with specific parameters (uniquely defining state $S$), the set of object-model and action-model relationships $R$, such that $W^i_S = \{M^i, A^i, O^i, R^i\}$. 
    \par Let $W^i_{goal}$ denote a desired state of the world, and let $A\prime$ denote a chosen set of actions applied within $W^i_S$ resulting in $W^i_{S\prime}$, denoted by $A\prime(W^i_S) \Rightarrow W^i_{S\prime}$, with $t_{A\prime}$ denoting the time required to select $A\prime$, and function $f(W^i_{S},W^i_{goal}) \Rightarrow A\prime$ denoting planning (i.e., determining $A\prime$ from a world state and goal).
    \par Let $\frac{\partial x}{\partial D}$ denote the improvement on $x$ given new data or experience $D$ obtained by the system. Intelligence can then be defined as:

    \begin{equation}
    I_i = \alpha ||W^i_S - W^*_S||_w + \frac{\beta ||W^i_{goal} - (A\prime(W^i_S) \Rightarrow W^i_{S\prime})||_w}{\gamma  t_{A\prime}} + \delta \frac{\partial W^i_S}{\partial D} + \varepsilon \frac{\partial f(W^i_{S},W^i_{goal})}{\partial D} - \zeta \frac{\partial t_{A\prime}}{\partial D} 
    \end{equation}

    where the parameter vector $[\alpha,\beta,\gamma,\delta,\varepsilon,\zeta]$ governs the weights of each component of intelligence (it is possible to measure individual components by appropriately selecting these parameters). We use $||x - y||_w$ to denote the world-norm of the difference between $x$ and $y$, defined as:

    \begin{equation} 
    ||x - y||_w =   \frac{1}{1+\sqrt{m(M^x-M^y)^2+a(A^x-A^y)^2+o(O^x-O^y)^2+r(R^x-R^y)^2}}
    \end{equation}

     where constants $m$, $a$, $o$, and $r$, govern the  importance of world properties. Specifically, the terms in the equation represent:

    \begin{enumerate}
        \item $||W^i_S - W^*_S||_w$: quality of knowledge. How well the internal world-model represents the real world.
        \item $||W^i_{goal} - (A\prime(W^i_S) \Rightarrow W^i_{S\prime})||_w$: quality of planning. How close to the goal did the selected set of actions move the world state to.
        \item $t_{A\prime}$: performance of planning. How quickly the set of actions was selected.
        \item $\frac{\partial W^i_S}{\partial D}$: world model learning. How much more comprehensive the internal world model is, given new data and experience.
        \item $\frac{\partial f(W^i_{S},W^i_{goal})}{\partial D}$: planning learning. How much better internal reasoning is at selecting actions to achieve a goal, given new data and experience.
        \item $\frac{\partial t_{A\prime}}{\partial D}$: performance learning. How much faster the system is at selecting actions, given new data and experience.
    \end{enumerate}

\section{Discussion}\label{sec:discussion}

The proposed definition of intelligence, we argue, highlights a few key insights. 

\begin{itemize}
    \item There should not be a \textit{binary distinction of intelligence}; i.e., it is not about "being intelligent or not". Intelligence is continuous; this should come as no surprise, given that it is widely accepted that levels of intelligence vary across humans, but the binary reasoning still permeates discussions about AI.
    \item Intelligence is a combination of three properties.  Thus, any system that exhibits at least one of those properties is at least partially intelligent. LLMs should be considered partially intelligent: they demonstrate at least some knowledge of the world (at least for the world language), exhibit planning within that world (e.g., creating outputs that meet the requirements of prompts), and can learn new concepts that inform the previous two properties (at least within the short term). 
    \item Many properties that are not relevant to intelligence (e.g., agency) play a key role in the behavior of AI. Discussions on AI regulation, alignment, ethics, and so forth, should be pedantic about the use of these terms.
\end{itemize}

\par Our work faces a significant limitation: we have not described how to measure the terms that make up our intelligence equation. In fact, we believe this to be the biggest challenge we face towards measuring intelligence. Derivative terms are likely easily measured, using standard AB testing, once measuring linear terms is a solved problem. But it is unclear how to perform this measurement. First, any measurement is relative to the observer, since our knowledge of $W^*$ is incomplete, to say the least. Similarly, evaluating $A\prime(W^i_S) \Rightarrow W^i_{S\prime}$ is probably not feasible beyond comparison with the observer's created $A\prime$ (these are especially keen when it comes to AGI). Second, even for any domain-specific intelligence, defining boundaries between what should and should not be a part of the world model is fuzzy: for example, it is possible a chess-playing intelligence develops new connections between chess and previously unconnected branches of mathematics. If that connection is lost on the observer, world-model measurement will be biased.
\par In conclusion, we have provided a formalized, purely functional, definition of intelligence, and we have attempted to distinguish it from other properties that, whilst often bundled together, are not part of intelligence. This work can support other researchers towards effective and ethical development of AI and AGI. As a corollary to our analysis, it seems developing an objective, absolute test of intelligence is likely impossible. Developing tests of intelligence relative to observer intelligence is likely possible, and we hope this work informs that research.

\subsection*{Funding}
No funding to report for this work.

\subsection*{Conflicts of Interest}
The author(s) declare(s) that there is no conflict of interest regarding the publication of this article.

\subsection*{Data Availability}
No data to report for this work.

\bibliographystyle{apalike}
\bibliography{refs}

\end{document}